\documentclass{temp2/llncs}

\usepackage{multirow}
\usepackage{csquotes}
\usepackage{booktabs}
\usepackage{graphicx}
\usepackage{mathptmx} 
\usepackage{times} 
\usepackage{amsmath} 
\usepackage{amssymb}  

\usepackage{subfigure}
\usepackage{float}
\usepackage{booktabs}
\usepackage{mathalfa}
\usepackage[mathcal]{eucal}
\usepackage[scientific-notation=false]{siunitx}
\usepackage{color}
\usepackage[dvipsnames]{xcolor}

\usepackage[misc,geometry]{ifsym} 
\usepackage[hyperfootnotes=false,hyperfigures=false]{hyperref}

\renewcommand{\thefootnote}{\fnsymbol{footnote}}

\title{Automatic brain tumor grading from MRI data using convolutional neural networks and quality assessment}

\author{S\'{e}rgio Pereira\inst{1,2}\textsuperscript{\Letter} \and Raphael Meier\inst{3} \and Victor Alves\inst{2} \and Mauricio Reyes\inst{4} \and Carlos A. Silva\inst{1}\textsuperscript{\Letter}}

\institute{CMEMS-UMinho Research Unit, University of Minho, Guimar\~{a}es, Portugal \\ \email{id5692@alunos.uminho.pt}; \email{csilva@dei.uminho.pt} \and Centro Algoritmi, University of Minho, Braga, Portugal \and Support Center for Advanced Neuroimaging, Institute for Diagnostic and Interventional Neuroradiology, University Hospital Inselspital and University of Bern \and Institute for Surgical Technology and Biomechanics, University of Bern, Switzerland}

\begin{document}

\maketitle

\renewcommand*{\thefootnote}{\arabic{footnote}}
\setcounter{footnote}{0}

\begin{abstract}

Glioblastoma Multiforme is a high grade, very aggressive, brain tumor, with patients having a poor prognosis. Lower grade gliomas are less aggressive, but they can evolve into higher grade tumors over time. Patient management and treatment can vary considerably with tumor grade, ranging from tumor resection followed by a combined radio- and chemotherapy to a ``wait and see'' approach. Hence, tumor grading is important for adequate treatment planning and monitoring. The gold standard for tumor grading relies on histopathological diagnosis of biopsy specimens. However, this procedure is invasive, time consuming, and prone to sampling error. Given these disadvantages, automatic tumor grading from widely used MRI protocols would be clinically important, as a way to expedite treatment planning and assessment of tumor evolution. In this paper, we propose to use Convolutional Neural Networks for predicting tumor grade directly from imaging data. In this way, we overcome the need for expert annotations of regions of interest. We evaluate two prediction approaches: from the whole brain, and from an automatically defined tumor region. Finally, we employ interpretability methodologies as a quality assurance stage to check if the method is using image regions indicative of tumor grade for classification.

\end{abstract}

\section{Introduction}

Gliomas are the most common primary brain tumors, being graded according to their malignancy. The most aggressive one is Glioblastoma Multiforme (GBM). These high grade gliomas (HGG) proliferate and infiltrate the surrounding tissues at a very fast pace. In fact, patients have a very short life expectancy, even if under treatment \cite{van2010exciting}. Lower grade gliomas (LGG) are less aggressive, and patients have a better prognosis. Nevertheless, LGG can evolve into HGG, hence, follow-up is required \cite{grier2006low}. Glioma grading is crucial when deciding the treatment procedure, which can range from surgery followed by chemo- and radiotherapy, to a ``wait and see'' approach. The latter avoids invasive procedures and is more common with LGG \cite{grier2006low,menze2015multimodal}. 

Histopathological diagnosis of biopsy specimens is the gold standard for glioma grading. However, it is time consuming, invasive, and prone to sampling error \cite{zacharaki2009classification}. MRI is the standard imaging technique for brain tumor diagnosis in clinical practice. In general, attributes of HGG in MRI include the contrast enhancing tumor tissue, necrotic core, edema, non-enhancing tumor, and mass effect. LGG are usually more diffuse, non-enhancing, smaller, and cause less mass effect. Nonetheless, some HGG may have some attributes of LGG, and vice-versa \cite{grier2006low,steed2018quantification,van2010exciting}. Tumor grading from imaging data would be useful in clinical practice, since it would avoid the sampling error, and expedite treatment planning by anticipating the histopathological results \cite{zacharaki2009classification}. Additionally, it would avoid the invasive biopsy procedures during follow-up. Studies suggest that perfusion MRI is more informative for glioma grading than structural MRI sequences \cite{zacharaki2009classification}. Still, perfusion MRI is not widely acquired in clinical practice \cite{essig2013perfusion}; in fact, perfusion MRI is seen as a plus, while structural MRI is part of the current consensus recommendations for standardized brain tumor imaging \cite{ellingson2015consensus}. Computer-based tumor grading from MRI is relatively unexplored. Zacharaki et al. \cite{zacharaki2009classification} predict the grade of gliomas from MRI images using a Support Vector Machine classifier. The method requires radiologists to manually define four regions of interest (ROI) in the tumor. Khawaldeh et al. \cite{khawaldeh2017noninvasive} use convolutional neural networks (CNN) in a semi-automated approach where the tumor grade is predicted from 2D slices selected by radiologists, which may result in multiple and possibly ambiguous predictions for the same patient.

CNNs offer the potential for learning tumor grading directly from imaging data without human-defined ROIs. However, these methods may fall into overfitting, and learn spurious patterns in the data. Hence, a quality assurance stage before deployment of these methods is desirable. As shown by Pereira et al. \cite{pereira2018enhancing}, interpretability of machine learning methods, through explanations of their predictions, allows one to assess which parts of the MRI image are more important for a prediction. In this way, one can evaluate if a model is trustworthy. Moreover, explanations may provide hints on undesirable behaviors, and allow one to devise improving strategies. The contributions in this paper are the following.  i) We propose to use 3D CNN for automatic glioma grading from conventional multisequence MRI, either from the whole brain, or an automatically defined tumor ROI. ii) We assess  the predictions by means of visual explanations. In this way, we were able to assess the predictions' trustworthiness and, as shown in the experiments, detect a problem in pre-processing. Finally, iii) we validate our approach on a publicly available database, making it more easily comparable with future proposals.

\section{Methods}

The proposed grading system has two main stages: ROI extraction, and glioma grade prediction. Additionally, we have an interpretation of predictions stage that serves as prediction quality assessment, and we use it for two purposes. First, to evaluate if regions indicative of tumor grade are the most relevant ones for classification. Second, to identify possible problems with the method (e.g. focus on spurious patterns) and devise strategies to obtain better classifiers.

\subsection{Extraction of the region of interest}

We consider and evaluate glioma grading from two ROI: the whole brain, and the tumor region. First, we automatically identify these regions in the image, and define a bounding box around them. Second, these volumes are extracted, resized to a fixed size, and fed into the tumor grade classification CNN. We note that an independent CNN is trained for each of the ROI. Regarding the whole brain region, in a skull-stripped image a bounding box can be easily defined from the brain mask.

For the tumor ROI, a bounding box is defined after segmenting the whole tumor. In order to account for segmentation mistakes, we give a margin of 10 voxels in each side of the bounding box, while maintaining the aspect ratio of the tumor. Segmentation of the whole tumor from multisequence MRI is achieved with a 3D U-net-inspired \cite{ronneberger2015u} fully convolutional network; the network architecture is depicted in Fig. \ref{fig:grade} (top). A 3D patch is extracted from each MRI sequence, stacked as channels, and fed into the network. The encoder path is responsible for learning the higher order features. Max-pooling layers increase the field of view, but downsample the feature maps. Features computed by higher (deeper) convolutional layers are more abstract. However, these features  lack fine details that are important for segmentation. Since the feature maps are downsampled, we need to map the lower resolution feature maps back to the input patch resolution. This is done by upsampling. As we upsample feature maps, we sum them with the feature maps of equivalent size of lower layers of the encoder path. Further convolutional layers fuse the lower and higher level features. We also employ residual blocks with pre-activations \cite{he2016identity} that make training of deep networks easier. The last layer is a $1\times 1 \times 1$ convolutional layer, with sofmax activation.

\subsection{Glioma grading CNN}
\label{sec:grade}

\begin{figure}[htb]
\centering
\includegraphics[width=1.0\textwidth, keepaspectratio]{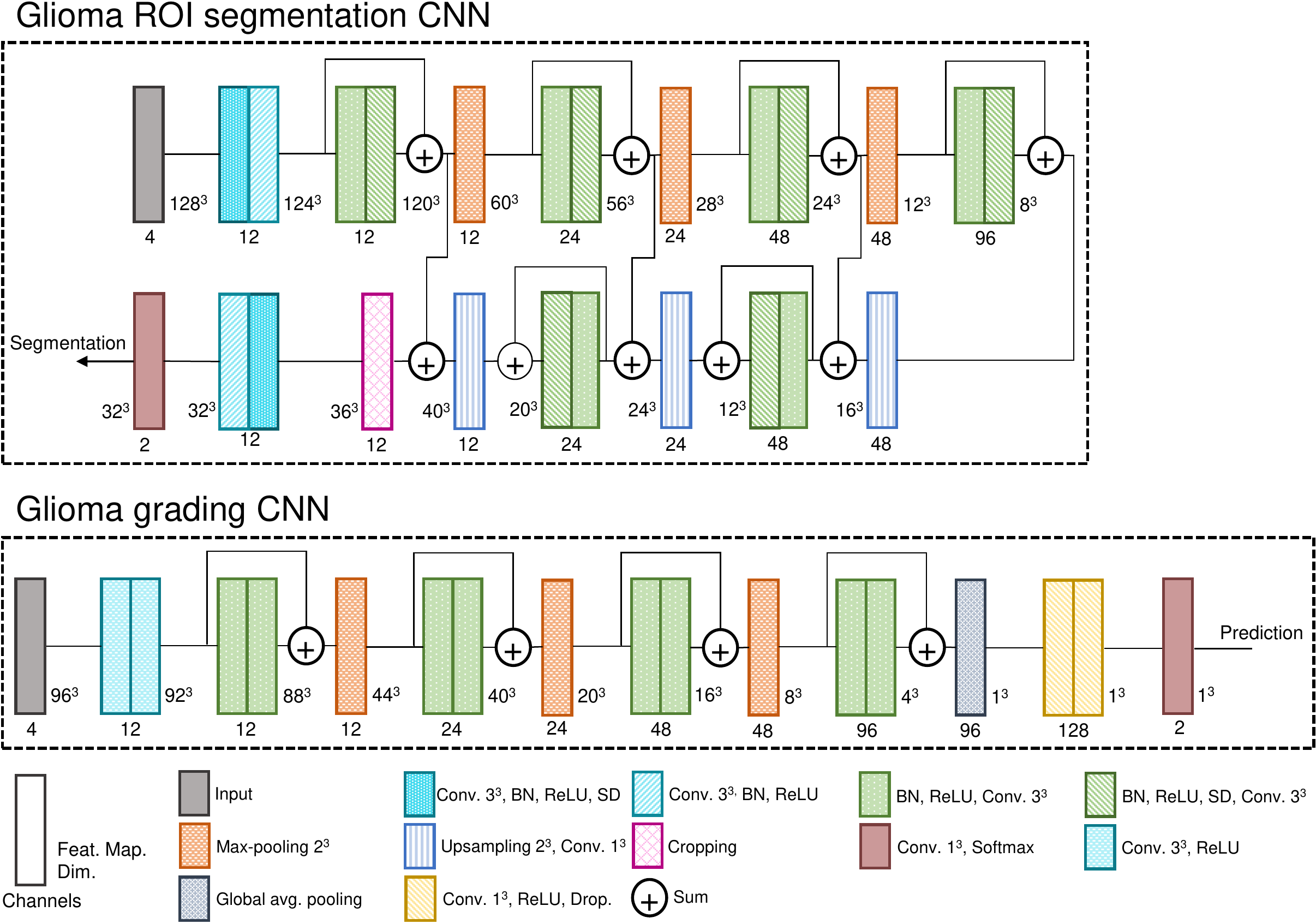}
\caption{Architectures of the CNNs used for glioma segmentation (top), and tumor grade classification (middle). Description of each block can be found in the bottom. BN stands for batch normalization, SD for spatial dropout \cite{tompson2015efficient}, and Drop. for dropout.}\label{fig:grade}
\end{figure}

We train a glioma grading CNN with similar architecture for each ROI (Fig. \ref{fig:grade}, middle). The ROI is extracted from each MRI sequence and resized to $96^3$, before feeding it to the CNN. In these architectures, we also employ residual convolutional blocks with pre-activations \cite{he2016identity}, which contribute for better learning. After the convolutional feature computation layers, we use Global Average Pooling to summarize each feature map. Then, a cascade of $1\times 1 \times 1$ convolutional layers act as fully-connected layers. Finally, the last layer outputs a probabilistic prediction of the tumor grade. Given the amount of available data, we use aggressive on-the-fly data augmentation during training. The data augmentation procedures were: sagittal flipping, rotation of [$\ang{-20}$, \ang{20}], \ang{90} rotation, and exponential intensity transformation with random $\gamma \in [0.85, 1.15]$.

\subsection{Grade prediction interpretability}

To perform quality assessment of tumor grade prediction, we use the interpretability methods Guided Backpropagation (GBP) \cite{springenberg2014striving} and Gradient-weighted Class Activation Mapping (GradCAM) \cite{gradcam}, after extending them to 3D. This is done at prediction time. 

Guided Backpropagation \cite{springenberg2014striving} is based on the idea that the gradient with respect to the input image, visualized in the image space, is informative of which parts of the image are more discriminative for the neurons activation. It starts by computing a forward pass through the network layers. During backpropagation, the true gradient is not calculated. Instead, a variation that results in better explanations of ReLU activations is used. This is performed by zeroing both the gradients in the units with 0 value after ReLU activation, and the negative gradients. In this way, the backward signals of neurons that contribute for decreased activation are discarded. Although visually discriminative, GBP has the disadvantage of not being discriminative in relation to the predicted class (i.e. it can highlight areas of interest to the network but not to which class). 

In contrast to GBP, GradCAM is class discriminative, but the explanation maps may have lower resolution. GradCAM tries to explain how the feature maps $F$ of a layer $l$ support the class prediction $y^c$. To that end, the gradient of the unit predicting the class with respect to the feature maps of the layer of interest $\frac{\partial y^c}{\partial F^l}$ is backpropagated. Then, the weight $\alpha_l$ of each feature map for the class prediction is computed as the global average pooling of the gradients. Being $i$, $j$, $k$ the indices of each of the N elements of the gradient, the weights are given by $\alpha{_l}^c = \frac{1}{N} \sum_{i} \sum_{j} \sum_{k} \frac{\partial y^c}{\partial F_{ijk}^l}$. Finally, the explanation map $E^c$ for the class is generated by the sum of $F^l$ weighted by $\alpha{_l}^c$, as $E^c = \max \left( \sum_l \alpha{_l}^c F^l, 0 \right)$. The $\max \left(\cdot, 0\right)$ function discards information contributing for decreased activation for the class. The explanation map has the same resolution as the feature maps of interest, thus, interpolation is typically needed to map results to the original image space.

\section{Experimental Setup}
\label{sec:exp_res}

The proposed methods were evaluated using BRATS 2017 Training set \cite{bakas2017advancing,menze2015multimodal}, which has the particularity that subjects are organized according to the tumor grade into HGG (GBM) and LGG. There are 285 pre-operative acquisitions: 210 HGG, and 75 LGG. For each subject there are 4 MRI sequences available with $1 \ mm$ isotropic resolution: T1, post-contrast T1 (T1c), T2, and FLAIR. All sequences are already aligned, and skull stripped. We randomly divided the 285 subjects into  60\% training, 20\% validation, and 20\% testing\footnote{Grades' proportions were maintained in each set. The subjects id in each set are available online: \url{https://github.com/sergiormpereira/brain_tumor_grading}.}. The manual segmentations of the different tumor compartments were merged into a single label to train the whole tumor segmentation network.

Two pre-processing steps are applied: bias field correction \cite{tustison2010n4itk}, and standardization of the image intensities inside the brain mask to zero mean and unit variance. All networks were trained with the Adam optimizer and crossentropy loss. For the whole tumor segmentation, learning rate (LR) was set to \num{5e-5}, spatial dropout probability to 0.05, and weight decay to \num{1e-6}. Regarding the CNNs for tumor grade prediction, the hyperparameters of the network were: LR -- \num{1e-4}, dropout probability -- 0.4, and weight decay -- \num{1e-4}. We used convolutional operations without padding, therefore, in skip connections, we cropped the feature maps to the same size of the smaller ones, before summing. During training, the bounding box of tumor ROI was defined using the manual segmentations. The grading CNNs were implemented with PyTorch and experiments were conducted using a NVIDIA GeForce Titan Black GPU.

For evaluation of tumor grading, we computed precision, recall, and f1-score. Since these metrics are influenced by class imbalance, we provide them for both LGG and HGG. Additionally, we compute the accuracy (acc) and the area under the receiver operating characteristic curve (ROC-AUC), which provide insights on the general ability of the classifier to distinguish between the classes. 

\section{Results and Discussion}
\label{sec:exp_res}

\begin{figure}[!tb]
\centering
\includegraphics[width=1.0\textwidth, keepaspectratio]{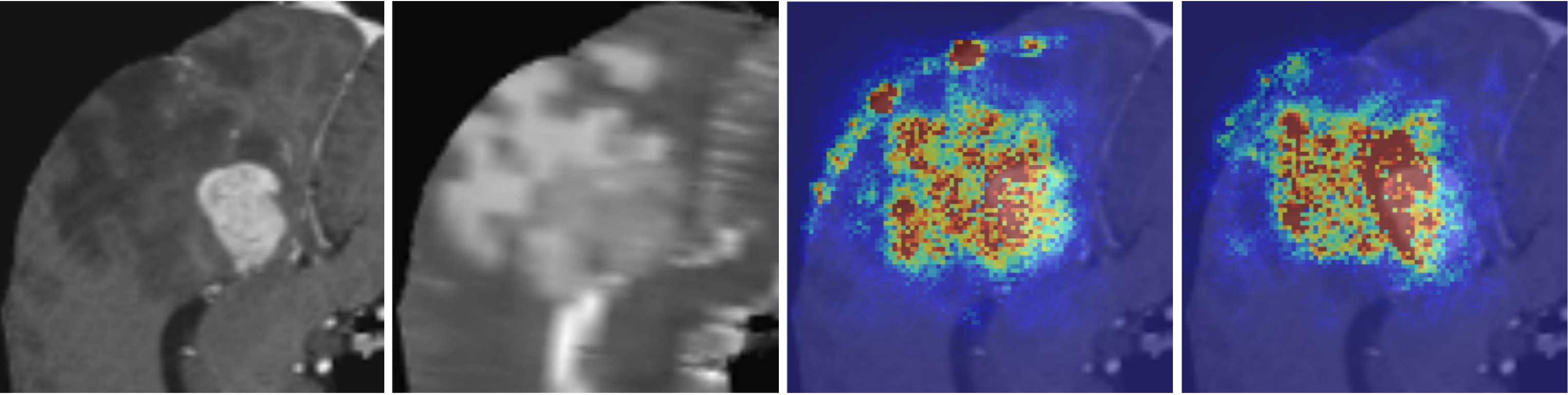}
\caption{Example of the effect of intensity standardization on the GBP maps. Warmer colors represent stronger responses. From left to right: T1c, T2, GBP map on image standardized over the whole image, and GBP map on image standardized in the brain region only.}\label{fig:preproc}
\end{figure}

Table \ref{tab:res} shows quantitative results for tumor grade prediction from each of the ROI (whole brain, and tumor). We note that it is expected to achieve lower f1-score, precision, and recall for LGG, since it is the minority class. Before feeding the images to the CNNs, we standardize the image intensities with zero mean and unit variance. Common approaches in the computer vision domain compute these statistics from the whole image. However, in MRI images, the background region is usually filled with 0 intensity values after skull stripping. When we standardize the intensities in the whole image, we achieve acc. of 0.895 (whole brain) and 0.877 (tumor ROI). However, from the GBP maps (Fig. \ref{fig:preproc}), we observe that the CNN considers the border of brain as discriminative, which for our data should not be a predictor of tumor grade. This is probably due to high gradients, since background has negative values, after standardization. Hence, we changed our pre-processing strategy by standardizing the image intensities inside the brain mask, only. After this approach, we observed that, mostly, the CNN does not consider the brain border as relevant for tumor grading. More interestingly, this simple change considerably boosted the metrics of tumor grade prediction from the tumor ROI (Table \ref{tab:res}). For instance, acc. and ROC-AUC improved from 0.877 and 0.8841 to 0.9298 and 0.9841, respectively. This shows an advantage of the interpretability stage, since it allowed us to identify a systematic problem and correct it; we note that the border problem would otherwise gone unnoticed, as results were already competitive.

\begin{table}[!tbp]
\centering
\caption{Tumor grade results for LGG and HGG in the two ROI: whole brain, and tumor. We show results for each variant of the image intensities standardization procedure.}
\setlength{\tabcolsep}{0.5em}
    \resizebox{1.0\textwidth}{!}{\begin{tabular}{lccccccc}
\specialrule{.2em}{.1em}{.1em}
 \textbf{Region} & \textbf{Standardization} & \textbf{Grade} & \textbf{F1-score} & \textbf{Precision} & \textbf{Recall} & \textbf{Acc} & \textbf{ROC-AUC} \\ \midrule 
\multirow{4}{*}{Whole brain} & \multirow{2}{*}{Whole image} & LGG & 0.8000 & 0.8000 & 0.8000 & \multirow{2}{*}{0.8950} & \multirow{2}{*}{0.8857} \\ 
 & & HGG & 0.929 & 0.929 & 0.929 &  &  \\ \cline{ 2- 8}
 & \multirow{2}{*}{Brain mask} & LGG & 0.8000 & 0.8000 & 0.8000 & \multirow{2}{*}{0.8950} & \multirow{2}{*}{0.8913} \\ 
 &  & HGG & 0.9286 & 0.9286 & 0.9286 &  &  \\  \midrule
\multirow{4}{*}{Tumor ROI} & \multirow{2}{*}{Whole image} & LGG & 0.7879 & 0.7222 & 0.8667 & \multirow{2}{*}{0.8770} & \multirow{2}{*}{0.8841} \\ 
 &  & HGG & 0.9136 & 0.9487 & 0.881 &  &  \\ \cline{ 2- 8}
 & \multirow{2}{*}{Brain mask} & LGG & 0.8667 & 0.8667 & 0.8667 & \multirow{2}{*}{0.9298} & \multirow{2}{*}{0.9841} \\ 
 &  & HGG & 0.9524 & 0.9524 & 0.9524 &  &  \\  \bottomrule
\end{tabular}}
\label{tab:res}
\end{table}

\begin{figure}[!htb]
\centering
\subfigure[]{\label{subfig:whole}\includegraphics[width=1.0\textwidth, keepaspectratio]{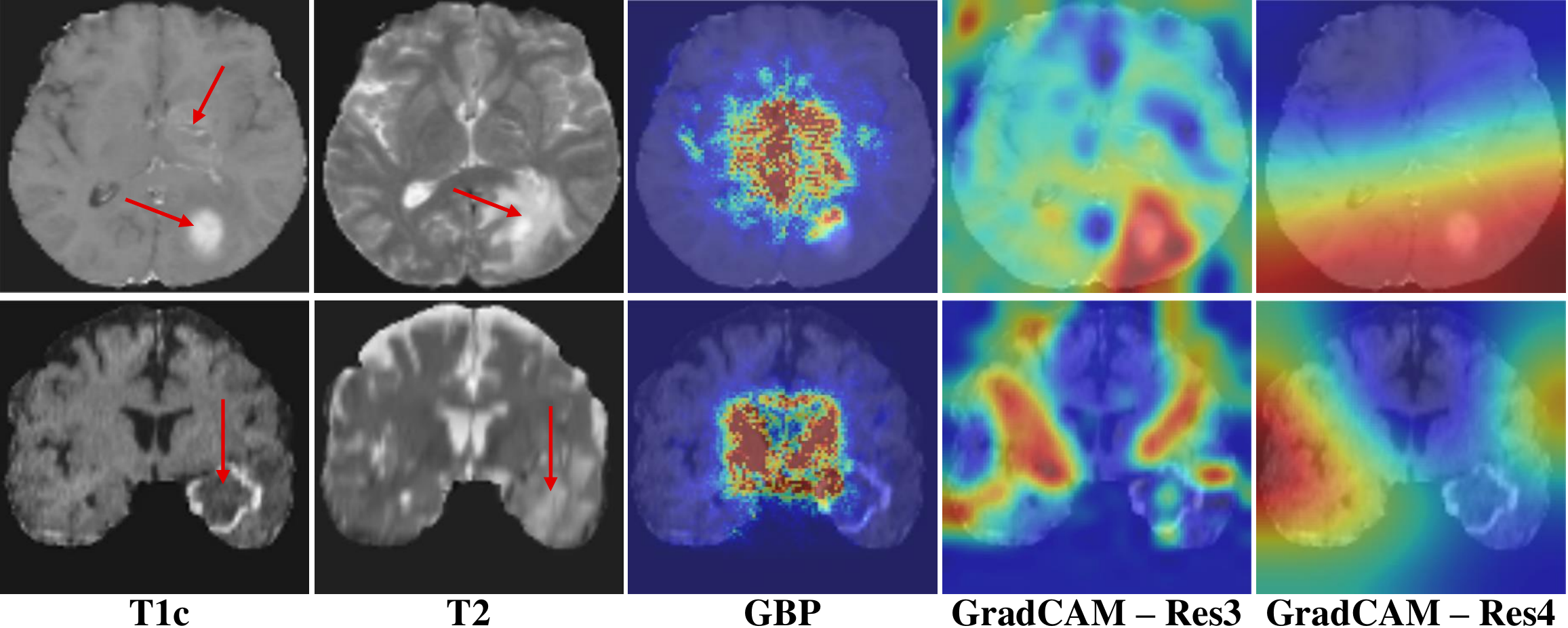}}
\subfigure[]{\label{subfig:roi}\includegraphics[width=1.0\textwidth, keepaspectratio]{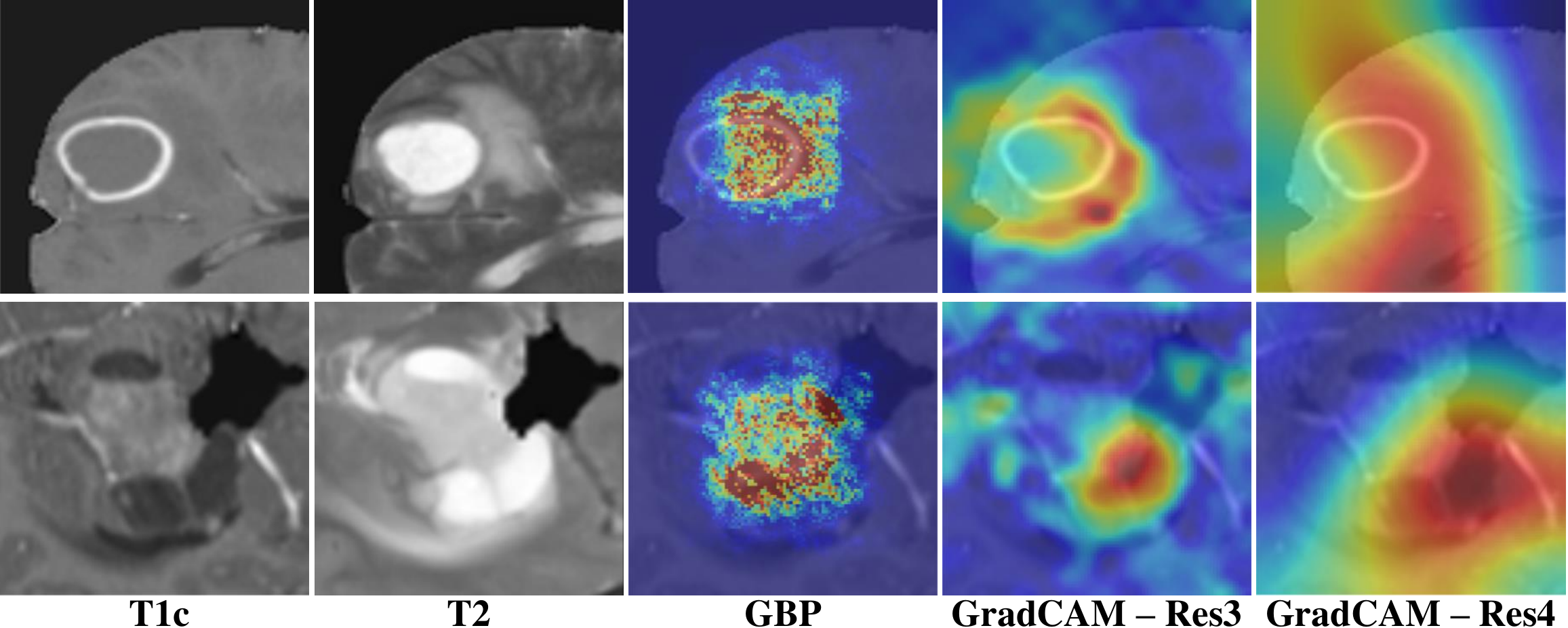}} 
\caption{Interpretability maps for grade predictions from a) whole brain, and b) tumor ROI. Warmer colors represent larger responses. In a) the arrows indicate the tumor lesions; on top is a correctly classified as HGG, while example in the bottom is a HGG misclassified as LGG. In b), the top example is a correctly classified HGG, while in the bottom a LGG is misclassified as HGG.}\label{fig:interp}
\end{figure}

Focusing on the variant with the standardization in the brain mask, we observe in Table \ref{tab:res} that grade prediction from the tumor ROI (acc -- 0.9298, ROC-AUC -- 0.9841) achieves better scores than grade prediction from the whole image (acc. -- 0.895, ROC-AUC -- 0.8913). Despite this, we note that tumor grade prediction from the whole brain achieves an acc. of 0.895, and f1-score of 0.9286, precision of 0.9286, and recall of 0.9286 for HGG. Fig. \ref{fig:interp} shows interpretability maps for some examples. We note that GradCAM provides maps with the same resolution as the feature maps of the layer of interest. We compute GradCAM maps with the output of the third (Res3) and fourth (Res4) residual blocks (Fig. \ref{fig:grade}). Fig. \ref{subfig:whole} shows interpretability maps for grade predictions from the whole brain. In the first row, the CNN was able to correctly grade it as HGG. From the two GradCAM maps we observe that the region of tumor was considered the most discriminative. The GBP shows focus on the ventricles, but, more interestingly, on both tumor lesions. In the second row, a HGG was mistakenly classified as LGG. The GradCAM maps are dispersed across the brain, instead of focusing in the tumor. We note that GradCAM is class discriminative, so, we show maps for LGG class. The GBP map concentrates in the ventricles. We observe that the CNN for tumor grading from the whole image focus on the ventricles frequently. We know that mass effect is a feature of HGG, and the ventricles are largely affected by it \cite{steed2018quantification}. Hence, the CNN may have learned that it is a predictor of malignancy. Actually, the subventricular zone is thought to be the origin of glioma cells, and nearby brain tumors are associated with worse prognosis \cite{liu2016anatomical}. The focus on ventricles may explain why the example in the second row is misclassified as LGG, since its effect on ventricles is smaller than the first row example. Fig. \ref{subfig:roi} shows examples of tumor prediction from the tumor ROI. In the first row, a HGG is correctly classified. From the GradCAM maps, we observe that the CNN correctly locates the tumor. Additionally, the Res3 and GBP maps appear to focus on the transition from necrosis to enhancing tumor and edema. This is in accordance with domain knowledge, as such an enhancing rim is characteristic for HGG. The second row of Fig. \ref{subfig:roi} is a LGG misclassified as HGG. In this case, it is a LGG with enhancing tumor. For this reason, the GradCAM maps for HGG and the GBP map seem to indicate that the enhancing tissues were responsible for the prediction, as it is a feature of HGG. It is possible that this is an evolving LGG that requires monitoring.

From the previous discussion, we see that GradCAM and GBP maps provide insights into the factors that contribute for a classification. So, we can see this interpretability stage as a quality assurance that enables us to check if the generated explanations are according to clinical knowledge. For instance, in the first row of Fig. \ref{subfig:whole} the explanations are focused on the tumor region. However, in the second row, the interpretability maps have high responses in regions that do not contain tumor. Thus, it may be a sign of an unreliable prediction, since it was based on regions of the image that are probably irrelevant. Additionally, the border effect problem, detected from the GBP maps, was a spurious pattern learned by the CNN.

\section{Conclusion}

Tumor grading from imaging data offers a fast and non-invasive approach for anticipating tumor grading, compared with histopathological diagnosis of biopsy specimens. We propose CNN for automatic brain tumor grading from MRI images, without the need of expert ROI definition. When we predict the grade from the whole brain, we achieve acc. of 0.895, while the prediction from the tumor ROI reaches an acc. of 0.9298. Therefore, our results show that grading is possible from both ROIs, although the latter achieves substantially better scores. Additionally, we employed interpretability approaches for prediction assessment, which allowed us to improve the pre-processing stage. Moreover, it may help in assessing if a decision is trustworthy by observing if it was actually based on the tumor region, or regions that are coherent with clinical knowledge. 

\subsubsection{Acknowledgments}

S\'ergio Pereira was supported by a scholarship from the Funda\c{c}\~ao para a Ci\^encia e Tecnologia (FCT), Portugal (scholarship number PD/BD/105803/2014). This work is supported by FCT with the reference project UID/EEA/04436/2013, COMPETE 2020 with the code POCI-01-0145-FEDER-006941.

\bibliographystyle{temp2/splncs03} 
\bibliography{references}

\begin{thebibliography}{10}
\providecommand{\url}[1]{\texttt{#1}}
\providecommand{\urlprefix}{URL }

\bibitem{bakas2017advancing}
Bakas, S., et~al.: Advancing the cancer genome atlas glioma mri collections
  with expert segmentation labels and radiomic features. Scientific data  4
  (2017)

\bibitem{ellingson2015consensus}
Ellingson, B.M., et~al.: Consensus recommendations for a standardized brain
  tumor imaging protocol in clinical trials. Neuro-oncology  17(9),  1188--1198
  (2015)

\bibitem{essig2013perfusion}
Essig, M., et~al.: Perfusion mri: the five most frequently asked technical
  questions. Am. J. Roentgenol.  200(1),  24--34 (2013)

\bibitem{grier2006low}
Grier, J.T., Batchelor, T.: Low-grade gliomas in adults. The oncologist  11(6),
   681--693 (2006)

\bibitem{he2016identity}
He, K., et~al.: Identity mappings in deep residual networks. In: ECCV. pp.
  630--645 (2016)

\bibitem{khawaldeh2017noninvasive}
Khawaldeh, S., et~al.: Noninvasive grading of glioma tumor using magnetic
  resonance imaging with convolutional neural networks. Applied Sciences  8(1),
  ~27 (2017)

\bibitem{liu2016anatomical}
Liu, S., et~al.: Anatomical involvement of the subventricular zone predicts
  poor survival outcome in low-grade astrocytomas. PloS one  11(4) (2016)

\bibitem{menze2015multimodal}
Menze, B.H., et~al.: The multimodal brain tumor image segmentation benchmark
  (brats). IEEE T Med Imaging  34(10) (2015)

\bibitem{pereira2018enhancing}
Pereira, S., et~al.: Enhancing interpretability of automatically extracted
  machine learning features: application to a rbm-random forest system on brain
  lesion segmentation. Med. Image Anal.  44,  228--244 (2018)

\bibitem{ronneberger2015u}
Ronneberger, O., et~al.: U-net: Convolutional networks for biomedical image
  segmentation. In: MICCAI. pp. 234--241. Springer (2015)

\bibitem{gradcam}
Selvaraju, R.R., et~al.: Grad-cam: Visual explanations from deep networks via
  gradient-based localization. In: ICCV (2017)

\bibitem{springenberg2014striving}
Springenberg, J.T., et~al.: Striving for simplicity: The all convolutional net.
  arXiv preprint arXiv:1412.6806  (2014)

\bibitem{steed2018quantification}
Steed, T.C., et~al.: Quantification of glioblastoma mass effect by lateral
  ventricle displacement. Sci.c Rep.  8(1),  2827 (2018)

\bibitem{tompson2015efficient}
Tompson, J., et~al.: Efficient object localization using convolutional
  networks. In: CVPR. pp. 648--656 (2015)

\bibitem{tustison2010n4itk}
Tustison, N.J., et~al.: N4itk: improved n3 bias correction. IEEE T. Med.
  Imaging  29(6) (2010)

\bibitem{van2010exciting}
Van~Meir, E.G., et~al.: Exciting new advances in neuro-oncology: The avenue to
  a cure for malignant glioma. CA: a cancer journal for clinicians  60(3),
  166--193 (2010)

\bibitem{zacharaki2009classification}
Zacharaki, E.I., et~al.: Classification of brain tumor type and grade using mri
  texture and shape in a machine learning scheme. Magn. Reson. Med.  62(6),
  1609--1618 (2009)

\end{thebibliography}

\end{document}